\documentclass[letterpaper, 10 pt, conference]{ieeeconf}  

\IEEEoverridecommandlockouts                              

\usepackage{graphicx} 
\usepackage{amsmath}      
\usepackage{mathtools}
\usepackage{caption}
\usepackage{subcaption}

\title{Learning Reach-Avoid Task with Reinforcement Learning: Vectorized Simulation and Benchmark}

\author{Jonas Weihing*, Shahram Eivazi*\\
\thanks{*Tübingen university}}

\begin{document}

\maketitle
\thispagestyle{empty}
\pagestyle{empty}

\begin{abstract}
Deep reinforcement learning (DRL) has a longstanding tradition in addressing the reach-avoid task problem, especially for controlling robotic arms. While this task serves as a baseline environment within the research community, the ability of DRL to effectively learn the reach-avoid task in complex and realistic scenarios beyond simplified and restricted tabletop settings remains uncertain. In this paper, we present, for the first time, a comprehensive benchmark for the reach-avoid task that accurately captures real-world complexities without simplifications. We demonstrate a diverse range of settings for robotic arm reach-avoid task, which can be used for evaluating DRL research. We achieved this by utilizing the MuJoCo MJX physics engine and parallelizing both the simulation environment and DRL algorithms using the Brax library. We achieved state-of-the-art results with success rates of 96.1\% (UR5e) and 98.8\% (Franka Emika Robot) for the reach task and 86.8\% (UR5e) and 95.2\% (Franka) for the static reach-avoid task. Our results indicate that while in previous works DRL agents could solve, for example, a reach task in a simplified setting perfectly, their agent’s performance collapses when evaluated in realistic scenarios. Overall, this work identifies that additional research is still required to claim the successful resolution of the robotic arm reach-avoid task using DRL. The environment and benchmarking code is available as open source at the following link (Blind Review).
\end{abstract}

\section{Introduction}
Recent advancements in robotic applications of deep reinforcement learning (DRL) have increasingly focused on precise object manipulation using robotic arms \cite{quillen2018deep, maurer2025synthesizing}. In fact, the success of these tasks largely depends on the DRL's ability to learn trajectories that reach a goal while avoiding any collisions within the environment. These trajectories are learned by DRL models either as sequences of cartesian positions for the robot arm's end-effector, which are then executed using classical inverse kinematics solvers \cite{lavalle1998rrt} or through an end-to-end approach, such as directly controlling robotic joints to achieve desired trajectories \cite{maurer2025synthesizing}.

In theory, DRL models should be able to learn both approaches for trajectory generation and performing precise object manipulation. However, addressing real-world complexities using DRL remains challenging due to the substantial amount of data required for training these models \cite{kalashnikov2018scalable}. 

Consequently, there is often a tendency to simplify tasks or restrict setups in DRL research, which can impact the evaluation of results \cite{jayawardana2022impact}. We believe that addressing these challenges is crucial for advancing the use of DRL in real-world scenarios. Surprisingly, this holds true even for fundamental DRL benchmarks and simulation tasks, such as OpenAI Gym \cite{brockman2016openaigym} environments or Panda-Gym \cite{gallouedec2021pandagym} robotic applications. 

This perspective on the practices and limitations of DRL in addressing real-world-like task settings serves as the foundation and driving force behind the work presented in this paper. 

With the rise of high-speed simulation software such as MuJoCo MJX \cite{todorov2012mujoco} and Nvidia Isaac \cite{nvidia_isaac_simulation}, which are highly used for their ability to synthesize large amounts of data rapidly, here, we introduce a comprehensive benchmark for a fundamental robotic task that accurately captures real-world complexities without simplifications.

\begin{figure}
    \centering
    \includegraphics[width=0.8\linewidth]{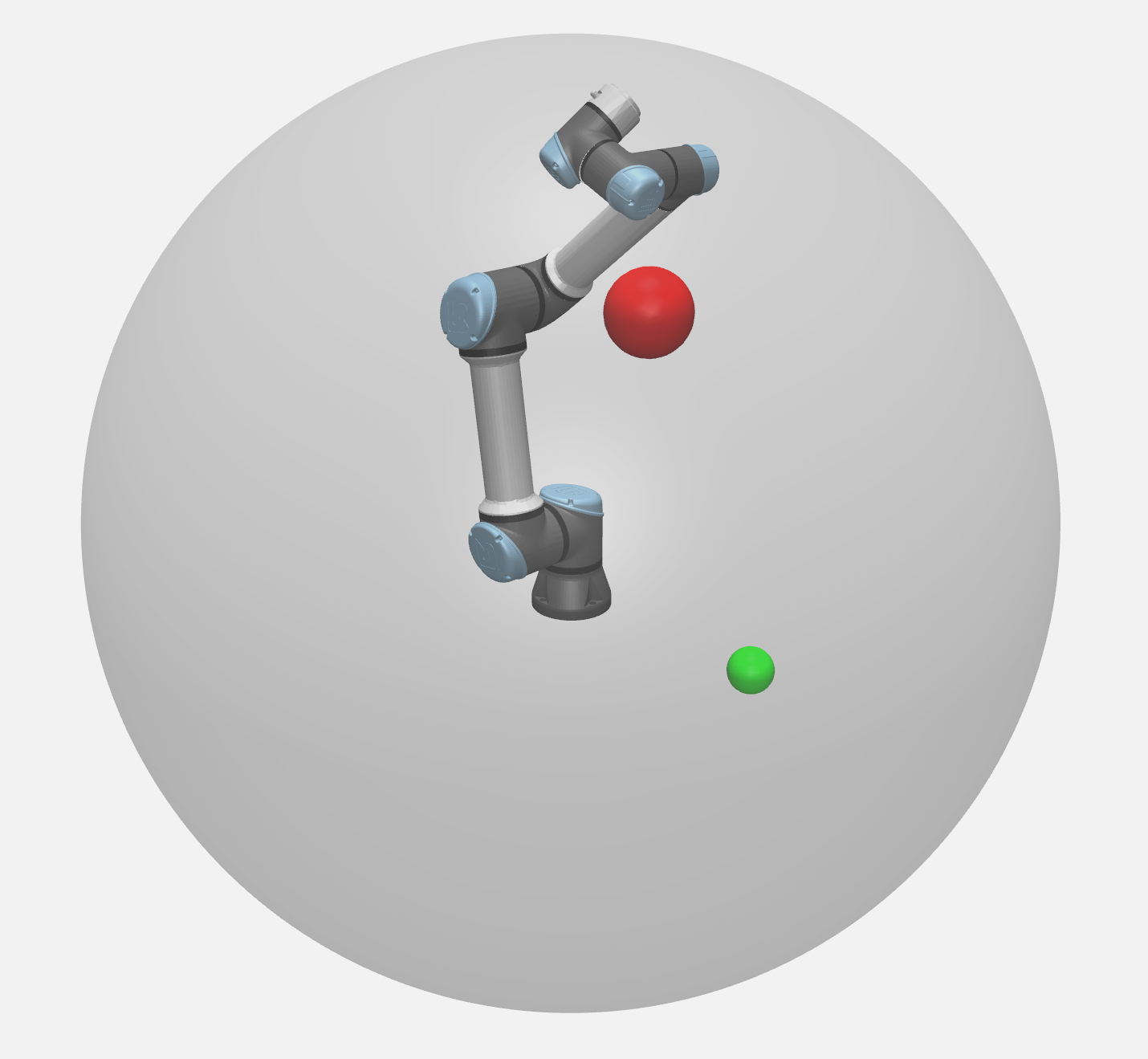}
    \caption{Visualization of a randomly sampled initial state for the reach-avoid task. The robot's full workspace is shown in gray, containing a randomly placed obstacle (red, 10 cm radius) and target (green). The target is visualized with an enlarged 5 cm radius for clarity.}
    \label{fig:ur5e_reach_pose_dist}
\end{figure}

Our focus is on developing a robot learning environment for the reach-avoid task in an end-to-end manner, operating in full working space with a DRL agent that directly predicts joint position targets. We used the Brax training pipeline alongside the MuJoCo MJX physics engine \cite{freeman2021brax} \cite{todorov2012mujoco}, to train Proximal Policy Optimization (PPO) \cite{schulman2017proximal} and Soft Actor-Critic (SAC) \cite{haarnoja2018soft} on massively parallelized simulations on a single GPU. We achieved over 10x speed-up in training compared to using stable baselines 3 \cite{raffin2021sb3} through the massive parallelization, running thousands of simulation instances simultaneously while keeping the entire DRL training loop in the same process on a single GPU. This minimizes the latency caused by CPU-GPU communication as well as inter-process communication, drastically reducing the time required for training and facilitating rapid experimentation. 

Our contributions are as followed:
\begin{itemize}
    \item We develop a reach-avoid task in a GPU-accelerated manner suitable for vectorized DRL training, enabling high-speed experimentation.
    \item We provide a comprehensive benchmark for the first time with a complex and realistic setting.
    \item We open source all environments and benchmarking code as open source.
\end{itemize}

\section{Related Work}
One fundamental and crucial objective in any robotic application is the reach-avoid task. This task can be viewed as an optimization problem, where the goal is to safely navigate a robot to a specified target and maintain its position there until a termination state is reached \cite{so2024solving}. Consequently, a substantial number of studies in the DRL literature focus on reach-avoid tasks, particularly in the context of articulated robotic arms \cite{pham2018optlayer, sangiovanni2018deep,luo2020accelerating,vasan2024revisiting}. This task involves long-horizon decision-making and complex motion sequences that require the end effector to reach the target pose while satisfying Cartesian space constraints and avoiding obstacles.

To solve the reach-avoid task, a common strategy among DRL studies is to substitute one or more components of classical path planning methods with DRL models. For instance, \cite{chiang2019rl,lindner2022reinforcement, bhuiyan2023deep} trained DRL agents that operate in end-effector Cartesian space. In these studies, the control of the robot joints is subsequently achieved using classical controllers and inverse-kinematic solvers. This Cartesian control approach is common in tasks such as pick-and-place and in benchmark environments like OpenAI Gym \cite{brockman2016openaigym} and Panda-Gym \cite{gallouedec2021pandagym}. Trained agents in such control modalities are ineffective in cluttered environments because the DRL models lack control over the full range of the robot's movements. For example, while the agent might learn to guide the end-effector around an obstacle, it cannot prevent collisions with the robot's elbow or other links.

To overcome the limitations of DRL-based Cartesian control, a growing number of researchers have focused on using DRL for direct joint-level control, whether through joint position, joint velocity, or torque commands \cite{pham2018optlayer, sangiovanni2018deep, luo2020accelerating, vasan2024revisiting, gallouedec2021pandagym}. This approach allows for more precise control over the robot's full configuration space, making DRL fundamentally more capable of complex collision avoidance. 

All of these studies are conducted in constrained tabletop settings with small, limited workspaces. In such simplified contexts, the risks of self-collision or complex obstacle avoidance are minimal, leading to a gap between these common benchmark results and the challenges of realistic scenarios. For example, \cite{wang2022ippo} observed that training with PPO \cite{schulman2017proximal} took an excessively long time, and the accuracy for the reaching task was poor, even without collision avoidance, when the workspace was increased to a quarter-spherical area.

The most related work that addresses this gap is Kumar et al. \cite{kumar2021joint}. They extended the robot's workspace to a more realistic size and included static obstacles. Their DRL model was able to map a Cartesian goal directly to a joint velocity control policy with a high success rate. However, their setup still represents only a subset of the robot's full working space. Furthermore, they reported being unable to train a PPO agent directly on the task, instead employing curriculum learning by gradually increasing the workspace in four stages. For the obstacle avoidance task, they reported performance comparable to that of reaching alone, but they did not provide a baseline for comparison, making it unclear how well the agent truly performed. Additionally, no information on self-collision and obstacle collision rates was provided, leaving open questions about the robustness of the learned policy.

In this paper, we integrate various aspects of previous works and present a detailed DRL benchmark for the reach-avoid task, utilizing the full workspace of a robotic arm. Our approach includes random start and goal positions, as well as obstacles, to comprehensively evaluate performance.

\section{Environment Setup}
To benchmark the ability of DRL to solve the reach-avoid task in complex, realistic scenarios, we developed a simulation framework using the MuJoCo MJX physics engine \cite{todorov2012mujoco}. The framework supports two widely used robots, the Universal Robots UR5e \cite{ur2025ur5e} and the Franka Emika Robot \cite{franka2025fer}. Our environment is built with a modular design that closely follows the Gymnasium API \cite{towers2024gymnasium} and is inspired by and seamlessly integrates with the Brax library \cite{freeman2021brax} for large-scale accelerated DRL training. 

Our implementation is based on a stateless design, which is essential for compatibility with JAX's just-in-time (JIT) compilation and automatic vectorization. All dynamic properties of the environment are encapsulated in a \textit{State} dataclass. This object is explicitly passed to and returned from pure functions like \textit{step} and \textit{reset}, enabling the entire simulation and training loop to run efficiently on a single GPU in a single jit-compiled process.
By mirroring the Brax interface, our custom environment can be used directly within the Brax training pipeline and its environment wrappers.

The software architecture is highly modular to ensure flexibility and extensibility. An abstract \textit{Env} class implements common, non-task-specific functionalities and defines the interface with the training pipeline. Specific tasks, such as reach and reach-avoid, inherit from this base class and implement its specific functionalities.
The robot specific logics, like the pose sampling or the sampling of a reachable position required for the target position, are encapsulated within a dedicated robot class with its own abstract base class, allowing different robots to be used interchangeably in the environments classes.
Higher-level logic, such as managing the episode length, is implemented in the Brax environment wrappers.

\subsection{Simulation settings}
The simulation parameters in the MJCF were carefully tuned for performance and stability. We used the default 2 ms timestep and the recommended Newton solver. To maximize speed, solver iterations were reduced to 1, line search iterations to 6, and the Jacobian was set to dense for optimal GPU execution. 
For the robots, we used the MJCF implementation from the MuJoCo Menagerie. With the UR5e, we replaced the end-effector's cylindrical collision geometry with a capsule, resulting in a 2x increase in simulation speed with minimal impact on collision accuracy. Additionally, the default controller parameters produced unrealistic joint velocities that exceeded the robot's real-world specifications by up to 5x. We corrected this by adding a joint damping force, which resulted in a more realistic velocity profile. 
For the Franka Emika Robot, we replaced the original mesh-based collision geometries with approximations with capsules, as they were incompatible with MuJoCo MJX.

\subsection{Simulation performance}
We benchmarked the environments sampling performance on a single NVIDIA RTX 3090 Ti GPU paired with an Intel Core i7-13700k. The number of simulation steps per second scales linearly with the number of parallel environments up to 8,192 environments. 
For the UR5e, the simple reach task achieved an average performance of approximately 185,000 steps per second with 8,192 parallel environments. The more complex reach-avoid task sustained a training throughput of around 125,000 steps per second. 
These figures are, however, directly dependent on the number of physics substeps (20 for this study) per environment step. For example, halving the substeps would roughly double the environment steps per second.

\section{Experiments}
We conducted our experiments on two independent tasks: 1) a foundational reach task in an obstacle-free environment, and 2) a more complex reach-avoid task requiring navigation around static obstacles (Figure \ref{fig:ur5e_reach_pose_dist}). 

\subsection{Reach Task}
In this task, the agent has to learn to manipulate the robot's joints to move its end-effector from a random initial configuration to a target position in its workspace. An episode is considered successful if the end-effector reaches the target without any self-collisions during the entire episode. 
For this, two task versions are common in research. A \textbf{reach-and-terminate} task ends the episode once the robot reaches the target. This may not ensure stable reaching, as a success might be due to luck only. The \textbf{reach-and-stay}, which is used in this study, continues the episode after a successful reach until collision or truncation. 

The episode length is set to 125 steps. With a simulation timestep of 2 ms and 20 physics substeps per environment step, this corresponds to 5 seconds in simulated time. The agent typically reaches the target in 20 to 50 steps, requiring the agent to learn a stable reach to remain in the target for the rest of the episode.

\subsubsection{Action Space}
We evaluated two joint control schemes: \textbf{absolute position control}, where the agent outputs the target joint angles directly, and \textbf{relative position control}, where the agent has to output deltas that are added to the current joint positions.

\subsubsection{Observation Space}
The observation space is designed to be configurable. It includes the most essential state information: the robot's current joint positions and the Cartesian position of both the end-effector and target. To provide the agent with information about the system's dynamics, the observation can be extended with  joint velocities and accelerations as well as the end-effector's Cartesian velocity.
To facilitate learning, the vector from the end-effector to the target and the action taken in the previous step are also included.

\subsubsection{Reward formulation}
We evaluated three common reward formulations. The first reward is a simple \textbf{sparse reward}, where the agent receives a reward of \(+1\) if the target is reached successfully and 0 otherwise. Collisions are penalized implicitly by terminating the episode, which prevents the agent from accumulating future rewards. If the target is reached and a collision occurs in the same environment step, the target is considered to be not reached, and the reward for this step is 0.
The second \textbf{semi-sparse reward} extends the sparse formulation with a linear distance-based term to encourage the agent to progress towards the target, as defined in Equation \ref{eq:reaching_semisparse_reward}.
\begin{equation}
    \begin{aligned}
        r_{semi-sparse} = {}&  
            w_{reach} \cdot
            \begin{cases}
                1, & \text{if target reached} \\
                0, & \text{else} \\
            \end{cases}
            \\ 
             &+ w_{dist} \cdot (2 \cdot d_{max} - d_{target})
        \label{eq:reaching_semisparse_reward}
    \end{aligned}
\end{equation}
Here, \(d_{max}\) is the robot's maximum possible reach, \(d_{target}\) the Euclidean distance from the end-effector to the target, and \(w_{reach}\) and \(w_{dist}\) are weighting factors.
This formulation creates a positive reward that increases as the agent approaches the target. By subtracting the distance from twice the robot's reach, the reward remains non-negative in almost all cases.
The third formulation is a \textbf{dense reward}, which uses an exponential function for the distance to the target, providing a strong learning signal when the end-effector is close to the target (Equation \ref{eq:reaching_dense_reward}).
\begin{equation}
    r_{dense} = 
        w_{dist} \cdot \exp(-\alpha \cdot d_{target})
    \label{eq:reaching_dense_reward}
\end{equation}
Here, \(\alpha\) is configurable to control the sharpness of the exponential curve and is set to 20.
Unlike the linear semi-sparse reward, this does not give an additional reward bonus if the target is reached and therefore is effectively independent of the target size. Through the reward signal as the agent nears the target, high-precision reaching is encouraged.

Additionally, for the semi-sparse and dense rewards, penalties on the joint velocities and accelerations can be added to penalize energy consumption and encourage smoother, more moderate movements. These are calculated based on the L2 norm of the joint velocity and acceleration vectors, respectively.

\subsection{Reach-Avoid Task}
The reach-avoid task extends the reach task into a multi-objective problem where the agent must balance minimizing the distance to the target while maximizing the distance to the obstacle. The action space is identical to the reach task, but the observation space includes additionally the obstacle's Cartesian position and its dimensions (represented as a 3D vector). 
The reward function builds upon the semi-sparse and dense formulations by adding a penalty term that discourages proximity of the robot's links to the obstacle, as shown in Equation \ref{eq:obstacle_reward} and adapted from \cite{kumar2021joint}.
\begin{equation}
    \begin{aligned}
        \mathbf{d_i} &= \| \mathbf{p}_{link,i} - \mathbf{p}_{obst} \|_2 - s_{obst}
        \\
        r_{obst} &= w_{obst} \cdot \sum_{i=0}^{n} \max \left(0, 1 - \frac{\mathbf{d}_{i}}{\tau}\right)
        \label{eq:obstacle_reward}
    \end{aligned}
\end{equation}
Here, for each of the \(n\) robot links, \(\mathbf{d_i}\) is the distance from the link's Cartesian position to the obstacle position, minus the size of the obstacle \(s_{obst}\) in the direction of the link. The penalty is applied if this distance falls below a configurable threshold \(\tau\).

\subsection{Training and evaluation}
All experiments were conducted using the JAX implementations of Soft Actor-Critic (SAC) \cite{haarnoja2018soft} and Proximal Policy Optimization (PPO) \cite{schulman2017proximal} from Brax. Training was performed on an Nvidia RTX 3090 Ti GPU, with each experiment run across 5 random seeds. The final performance was averaged over these seeds, evaluating each on 4096 different environments. 
For PPO, we used a batch size of 512, 64 minibatches, a learning rate of \(8\cdot10^{-4}\), and a discount factor \(\gamma\) of \(0.97\). The number of updates per batch was set to 5 or 7, depending on the control scheme. The unroll length is set to 10 and no action repeat is used. The network architecture for both policy and value functions was set to three hidden layers with 256 neurons each. 
For SAC, we used 128 parallel environments, a batch size of 256, a learning rate of \(1\cdot10^{-4}\), and a reward scaling of 20. The discount factor and action repeat are kept the same. The size of the replay buffer was set to a minimum of 102,400 and a maximum of 2,048,000 transitions.

Performance is measured using two primary metrics: the target reached rate and the collision rate, which is further subdivided into self-collisions and obstacle collisions. An episode is marked as “target reached” if the target is reached with the end-effector at least once without any collision during the episode. To assess final accuracy, we also report the average end-effector distance to the target over the last 10 steps of the episode. 

\subsubsection{Initial pose and target sampling}
At the beginning of each episode, the robot's initial joint configuration and the target's Cartesian position are sampled from uniform distributions. Initial poses are drawn from the robot's full joint range, rejecting those that lead into a collision. The targets are sampled either uniformly from the spherical workspace or from the Cartesian end-effector position of a randomly sampled pose of the robot, ensuring the target is reachable. The latter, however, leads to a non-uniform distribution, as seen in Figure \ref{fig:ur5e_reach_pose_dist}.
\begin{figure}
    \centering
    \includegraphics[width=1.0\linewidth]{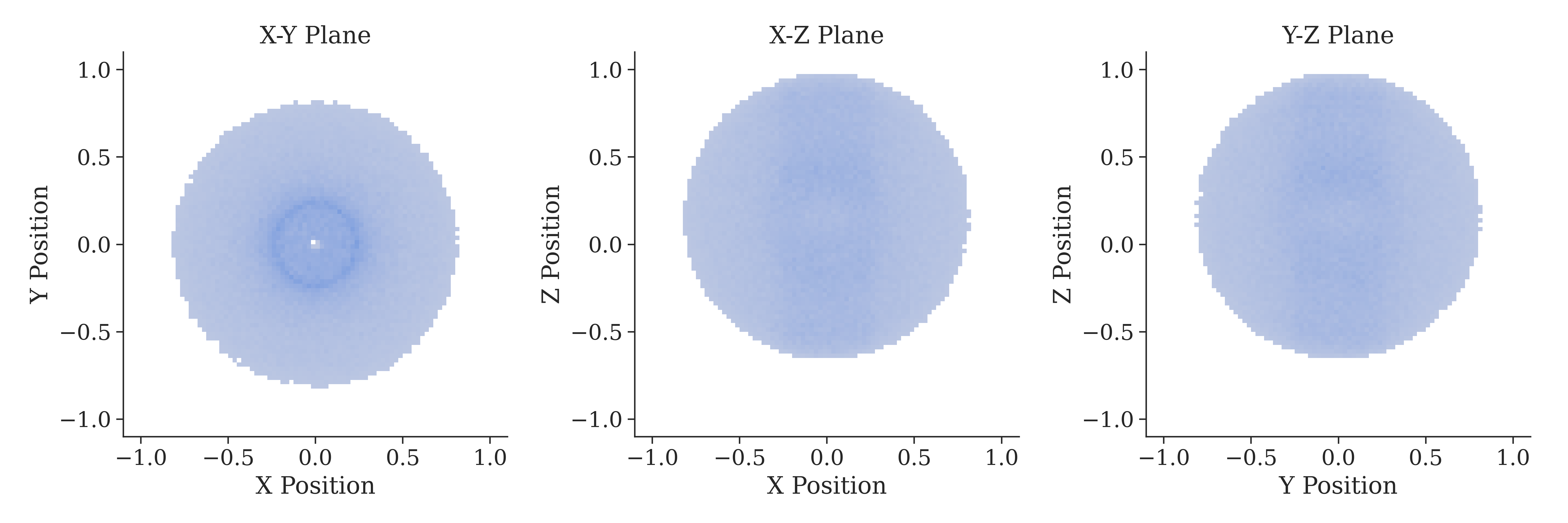}
    \caption{Distribution of reachable target positions for the UR5e when sampling from the end-effector of a uniformly sampled, collision-free pose.}
    \label{fig:ur5e_reach_pose_dist}
\end{figure}
The obstacles were sampled uniformly in a 1.2m×1.2m×1.2m cube, centered around the world origin, rejecting positions that result in a collision upon initialization.

\subsubsection{Baselines}

For the reach task, we compared against three baselines: a random action agent, a standstill that uses its initial pose as the action, and a pseudo inverse kinematics (IK) agent that sets the action to a joint position that is known to be able to reach the target. The results are shown in Table \ref{tab:reaching_baselines}.
\begin{table}
    \centering
    \caption{Performance of the baselines on the reaching task for two different target sizes (1 and 2 cm). The performances are reported in rates between 0 and 1, where 1 would be equivalent to 100\%.}
    \begin{tabular}{ccccc}
        \hline\hline
        \textbf{Agent} & 
        \textbf{\begin{tabular}[c]{@{}c@{}}Target \\ size\end{tabular}} & 
        \textbf{\begin{tabular}[c]{@{}c@{}}Target \\ reached\end{tabular}} & 
        \textbf{\begin{tabular}[c]{@{}c@{}}Self \\ collision\end{tabular}} & 
        \textbf{\begin{tabular}[c]{@{}c@{}}Distance \\ to target\end{tabular}} \\
        \hline
        Random      & 0.01m  & 0.000 & 0.300 & 0.929m ± 0.328 \\
        Standstill  & 0.01m  & 0.000 & 0.000 & 0.855m ± 0.318 \\
        Pseudo IK   & 0.01m  & 0.369 & 0.222 & 0.033m ± 0.035 \\
        \hline
        Random      & 0.02m  & 0.001 & 0.300 & 0.929m ± 0.328 \\
        Standstill  & 0.02m  & 0.000 & 0.000 & 0.855m ± 0.318 \\
        Pseudo IK   & 0.02m  & 0.510 & 0.222 & 0.033m ± 0.035 \\
        \hline\hline
    \end{tabular}    
    \label{tab:reaching_baselines}
\end{table}
For the reach-avoid task, we used a reach agent as a baseline. When introduced to an environment with an obstacle, its target reached rate drops to 86.6\%, primarily due to obstacle collision, which occurred in 10.3\% of the episodes.

\section{Results}
The best-performing agent for the reach task in a full workspace achieved a \textbf{96.1\%} target reached with the UR5e (1 cm target size) and a 2.2\% self-collision. Results on the Franka Emika Padna were even better, reaching a \textbf{98.8\%} target reached rate with only 0.3\% self-collisions.
In contrast, SAC struggled with the UR5e, achieving only a 61.6\% target reached rate due to a high self-collision rate of 17.3\%. However, SAC performed better with the Franka Emika Panda, reaching 87.9\% target reached with 2.7\% self-collisions.
The increased self-collisions with the UR5e are mainly due to the accumulation of collisions between the wrist and forearm, which are physically impossible to collide on the Franka Emika Robot.
Halving the joint range from 720° to 360° for initialization on the UR5e leads to an increase in performance, with SAC reaching 84.8\% target reached and only 3.5\% self-collisions.
Given its superior performance and due to page limitations, we further use the PPO algorithm for all experiments unless otherwise noted.

\subsection{Action space}
We investigated the differences between the two action spaces, absolute position and relative position control, using a dense reward. While both achieved a statistically identical target reach rate of 96.5\% for absolute position and 96.4\% for relative position control, the absolute position control resulted in a 1.3 times higher self-collision rate (3.0\% vs. 2.3\%).
Furthermore, as shown in Table \ref{tab:reaching_actionspace_differences}, the control schemes differ significantly in precision and stability. The absolute position control agent struggles to maintain its pose, often jittering around the target due to outputting actions with large differences between consecutive steps. In contrast, the relative position control agent is far more precise and stable, remaining in the target for over twice as long and leaving it in only 5\% of the episodes at all, compared to an average 20 times per episode.
\begin{table}
    \centering
    \caption{Comparison of precision metrics between absolute position and relative position control. Distances are in meter. The \textit{Steps in target} and \textit{Times left target} metrics count how many steps of the episode the robot was in the target and how often it left the target.}
    \begin{tabular}{rccc}
        \hline\hline
        \textbf{\begin{tabular}[c]{@{}c@{}}Control \\ type\end{tabular}} & 
        \textbf{\begin{tabular}[c]{@{}c@{}}Distance \\ to target\end{tabular}} & 
        \textbf{\begin{tabular}[c]{@{}c@{}}Steps in \\ target\end{tabular}} & 
        \textbf{\begin{tabular}[c]{@{}c@{}}Times left \\ target\end{tabular}} \\
        \hline
        Position  & 0.012 ± 0.001   & 41.5 ± 3.2    & 21.7 ± 0.4    \\
        Relative  & 0.005 ± 0.000   & 99.1 ± 0.4    & 0.05 ± 0.00   \\
        \hline\hline
    \end{tabular}
    \label{tab:reaching_actionspace_differences}
\end{table}

\subsection{Reward formulations}
We further evaluated the three reward formulations: sparse, semi-sparse, and dense. Contrary to previous research suggesting the superiority of sparse reward \cite{vasan2024sparsereward}, our experiments show that a sparse reward agent fails to learn the reach task in a large workspace with a small target. 
This is due to the extreme difficulty of exploration, where the probability of a random policy reaching the 1 cm target is only 0.006\%, requiring an average of 2 million steps to receive the first non-zero reward signal.

\begin{figure}
    \centering
    \includegraphics[width=1.0\linewidth]{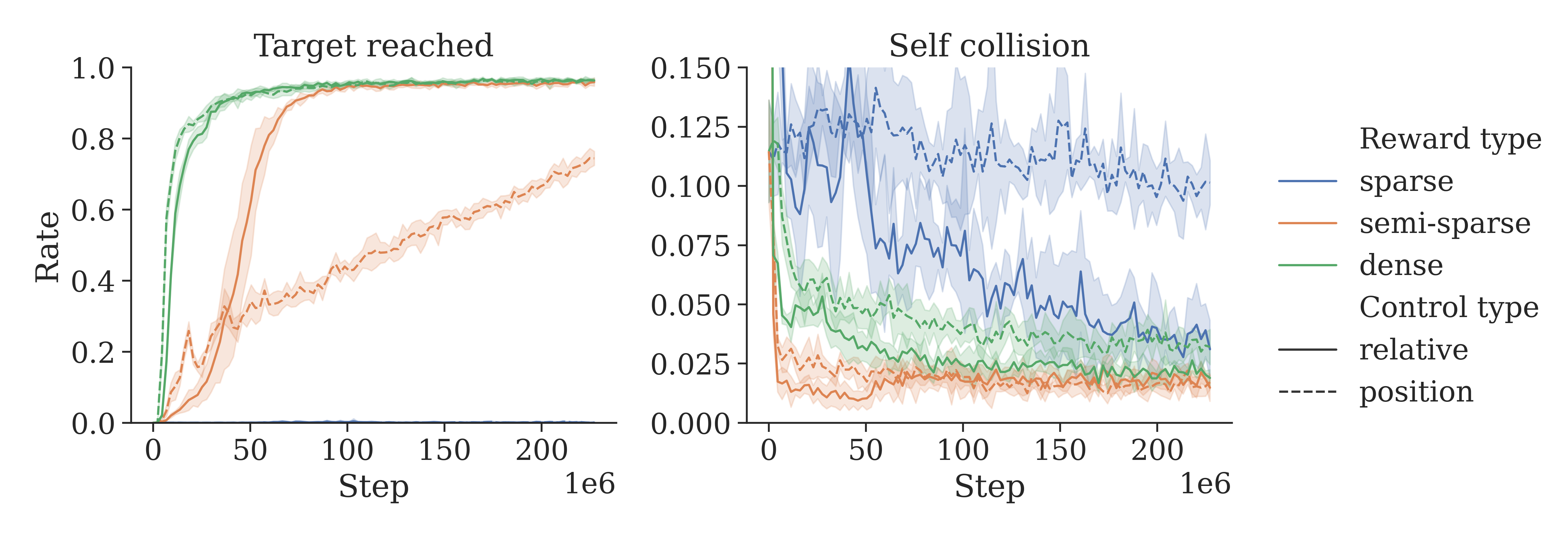}
    \caption{Comparison of different reward formulations. The dense reward learns significantly faster and achieves higher precision for both control schemes.}
    \label{fig:reaching_reward}
\end{figure}

Both the semi-sparse and dense rewards enabled successful learning. However, the dense reward agent learned faster and achieved a higher final precision with the relative position and especially the absolute position control, making it the preferred choice (Figure \ref{fig:reaching_reward}).

We also investigated adding penalties for joint velocity and acceleration. A small acceleration penalty of \(-0.0001\) reduced the average acceleration by up to 12\% without a statistically relevant influence on the overall performance but hindered initial learning. A larger penalty was counterproductive as the agent learned to collide immediately to terminate the episode and avoid accumulating negative rewards. Velocity penalties showed no significant benefit. Consequently, these penalties were omitted in all other experiments.

\subsection{Observation space}
The optimal observation space depends on the action space, and the experiments were conducted using the dense reward. For relative position control, a minimal observation of joint positions and the target position is theoretically sufficient. However, this configuration resulted in slow learning (150-200 million steps for 80\% target reached) and high variances between runs. Including the end-effectors Cartesian position significantly improved performance, leading to faster training, a higher target reached rate (96\%) and much greater precision (0.5 cm vs. 2-4 cm final distance to the target).
Increasing the network depth from 3 to 5 layers did not solve the minimal observations performance issues, suggesting it is not an under-capacity problem. Adding further information, such as velocities or the previous action offered no additional benefit.

For absolute position control, the best performance was achieved by including the end-effector position and previous action, as expected. Additional observations also provided no further improvements.

\subsection{Network size}
While DRL research often uses networks with 2-3 layers of 64-256 neurons \cite{schulman2017proximal, lua2020reaching}, we tested various larger architectures, due to the increased task complexity.
For relative position control, a standard architecture of 3 layers with 256 neurons each (3×256) for both policy and value networks proved sufficient. Increasing network size to 4 layers (4×256) or 3 layers with 512 neurons each (3×512), or using  larger first layers \cite{ota2021training}, like 1024 or 2048 neurons, yielded no benefits.
Reducing the network size to just two layers (e.g., to 2×128 or 2×256) proved to be insufficient and delayed the time to convergence. 

\begin{figure}
    \centering
    \includegraphics[width=1.0\linewidth]{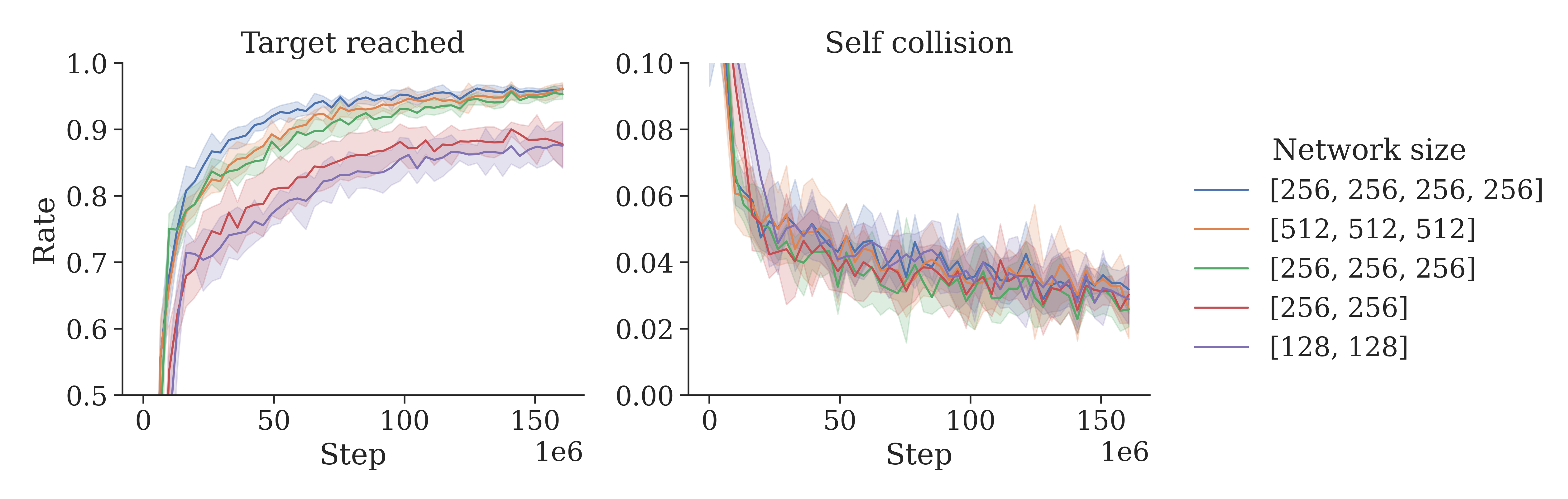}
    \caption{Comparison of different network sizes for position control. The \textit{4×256} architecture offers a performance increase to \textit{3×256} and a faster wall-clock time over \textit{3×512}.}
    \label{fig:reaching_position_network}
\end{figure}

For absolute position control, a 2-layer network was also insufficient over 3×256. The larger 4×256 architecture, however, offered a slight performance increase over the standard 3×256 and trained faster than a 3×512 with similar performance, making it more suitable (see Figure \ref{fig:reaching_position_network}).

\subsection{Influence of the workspace size}
\begin{figure}
     \centering
     \begin{subfigure}{0.2\textwidth}
         \centering
         \includegraphics[width=\linewidth]{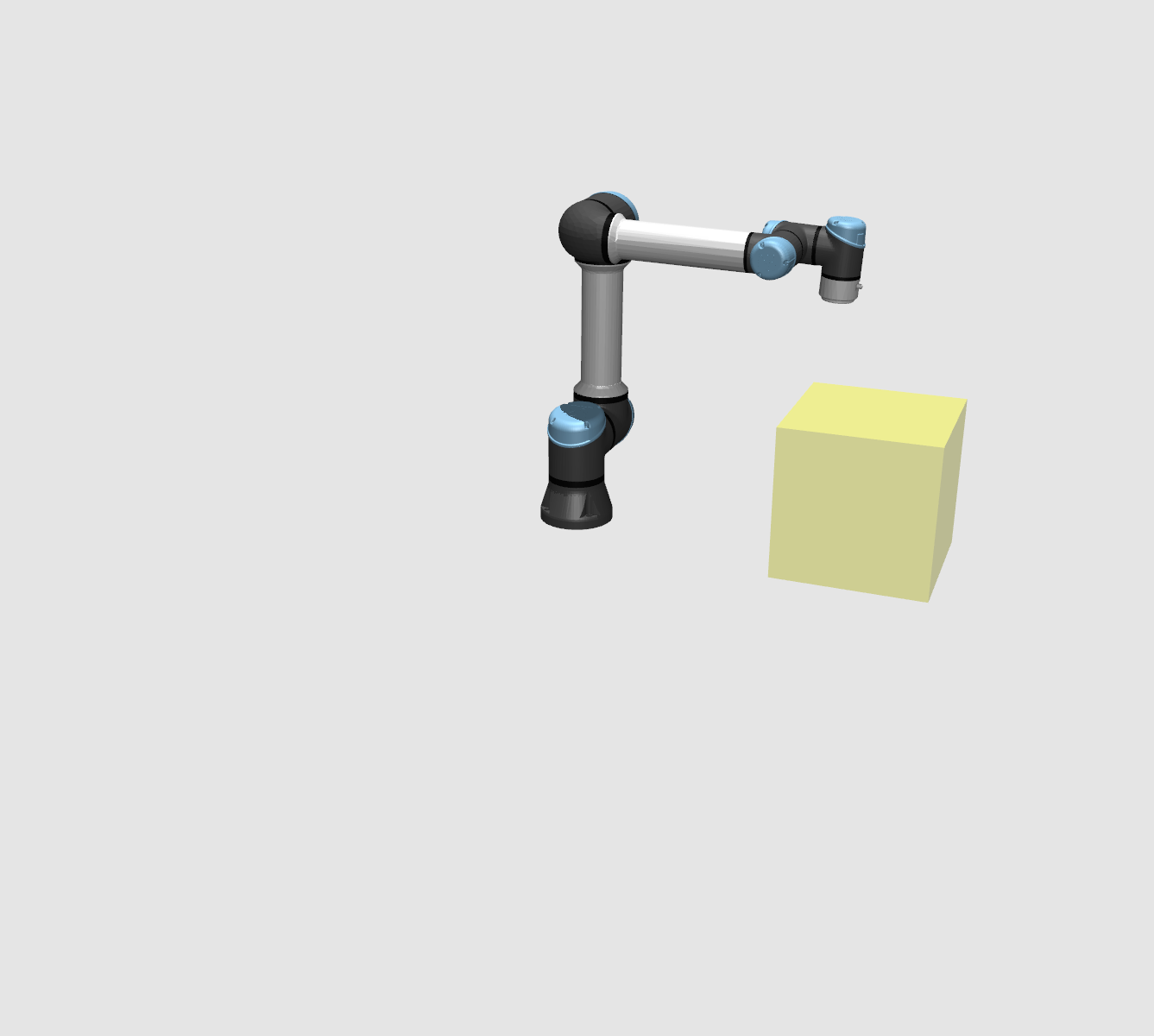}
         \caption{Pandagym}
         \label{fig:workspace_pandagym}
     \end{subfigure}
     \begin{subfigure}{0.2\textwidth}
         \centering
         \includegraphics[width=\linewidth]{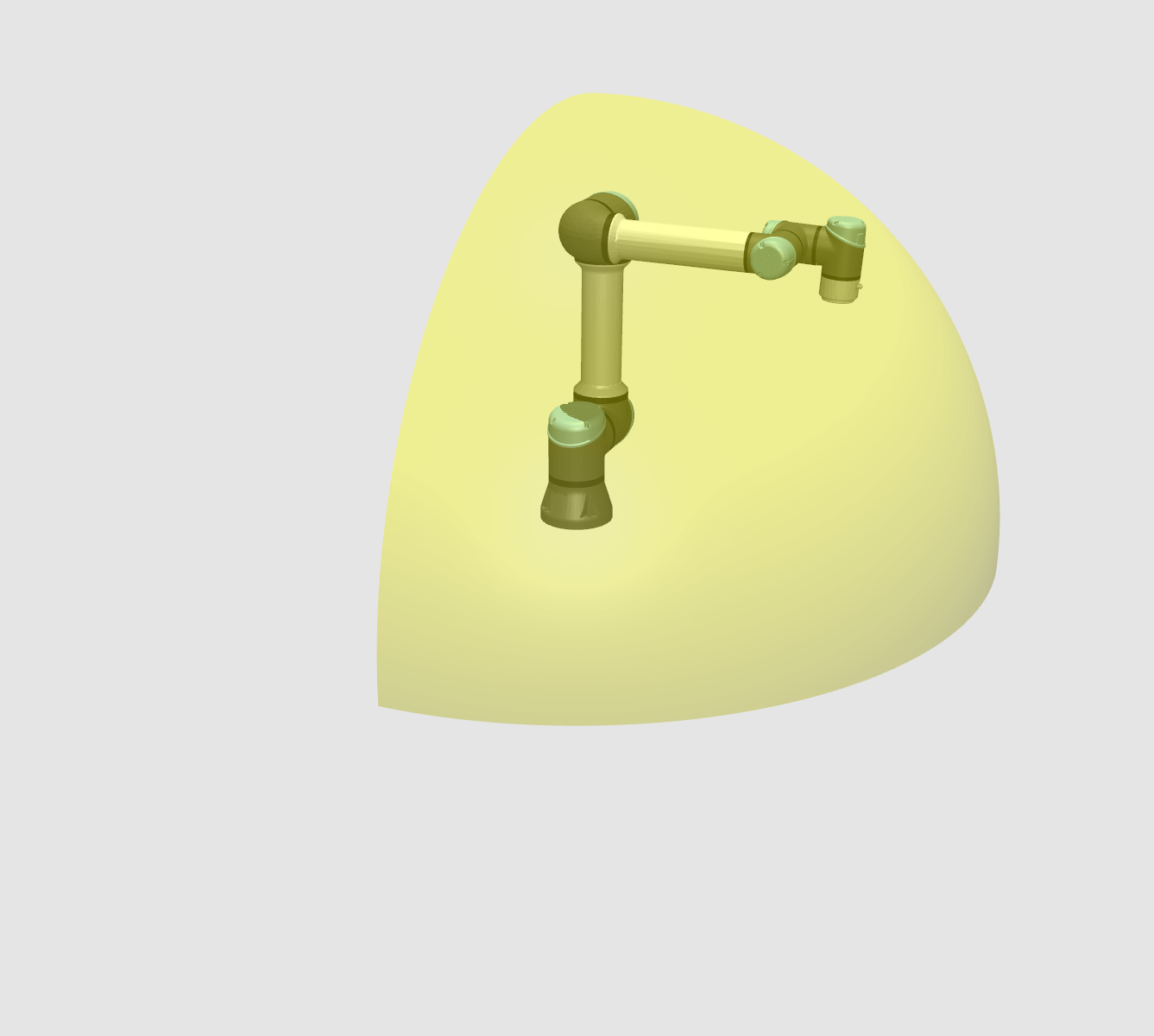}
         \caption{Front}
         \label{fig:workspace_front}
     \end{subfigure}
     \begin{subfigure}{0.2\textwidth}
         \centering
         \includegraphics[width=\linewidth]{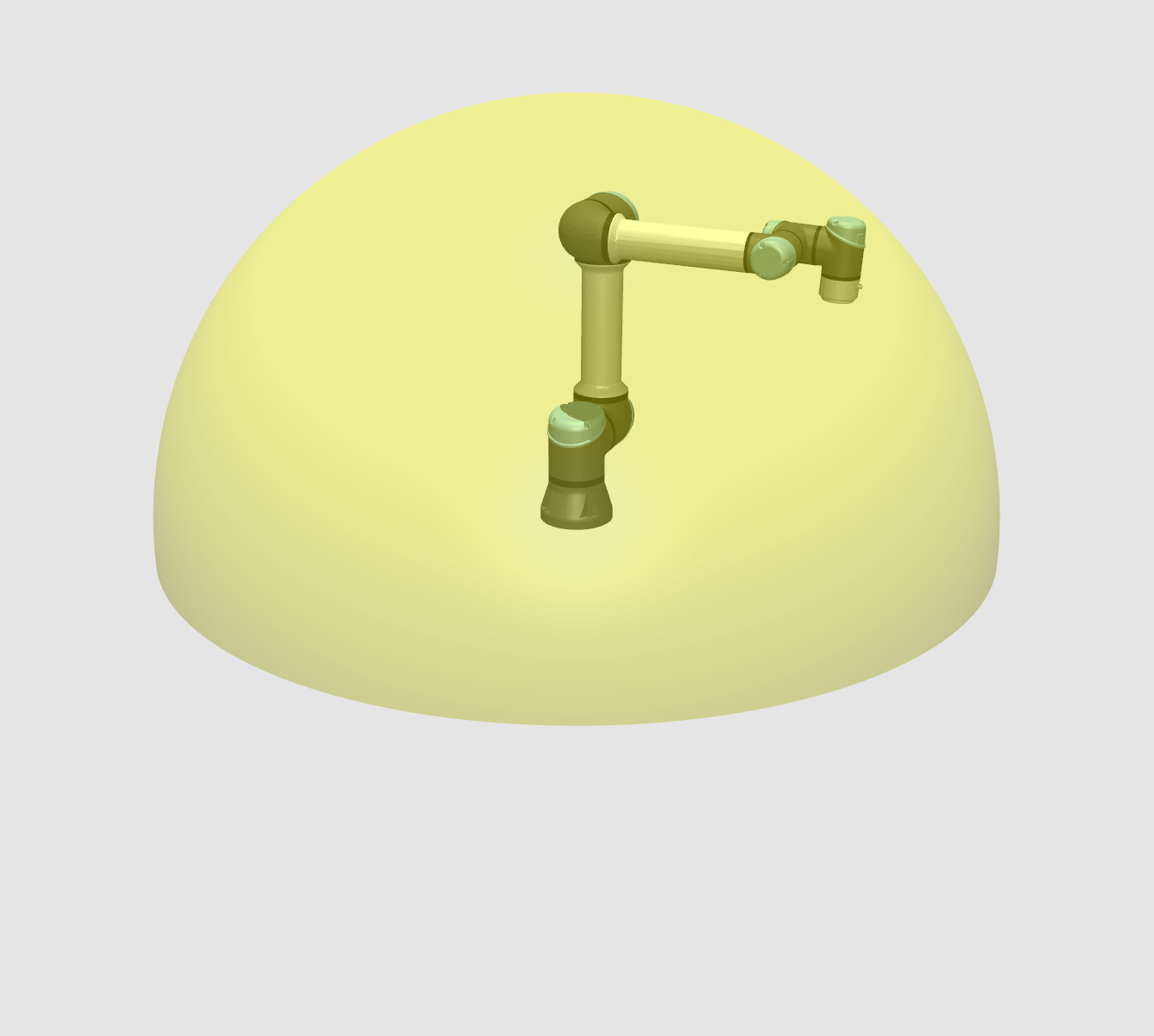}
         \caption{Upper}
         \label{fig:workspace_upper}
     \end{subfigure}
     \begin{subfigure}{0.2\textwidth}
         \centering
         \includegraphics[width=\linewidth]{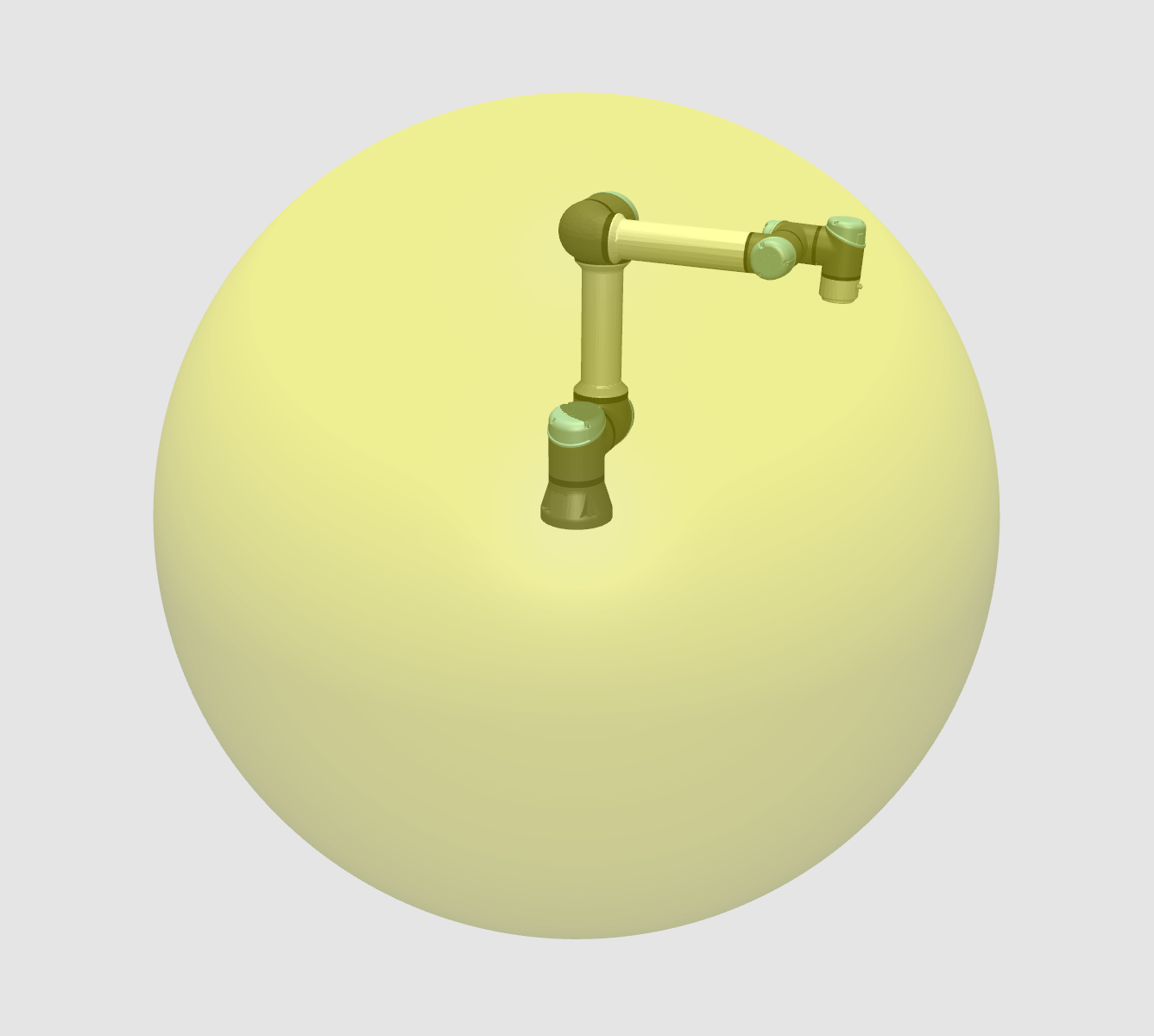}
         \caption{Full}
         \label{fig:workspace_full}
     \end{subfigure}
     \caption{Approximate visualization of the four different workspace settings used for our analysis. The \textit{PandaGym} \cite{gallouedec2021pandagym} workspace is a 30×30×30 cm cube located 45 cm in front of the robot. The \textit{Front} workspace is a 180-degree area in front of the robot. The \textit{Upper} workspace is a full tabletop setting. The \textit{Full} workspace is the entire reachable space of the UR5e.}
     \label{fig:workspace_size_comparison}
\end{figure}

Previous work on reach and reach-avoid tasks has often been limited to small, constrained workspaces like a tabletop setting \cite{kumar2021joint} or the 30×30×30 cm working area in PandaGym \cite{gallouedec2021pandagym} rather than the robot's full reachable area. We investigated how workspace size impacts policy learning across four distinct workspace settings, visualized in Figure \ref{fig:workspace_size_comparison}.

\begin{figure}
    \centering
    \includegraphics[width=1.0\linewidth]{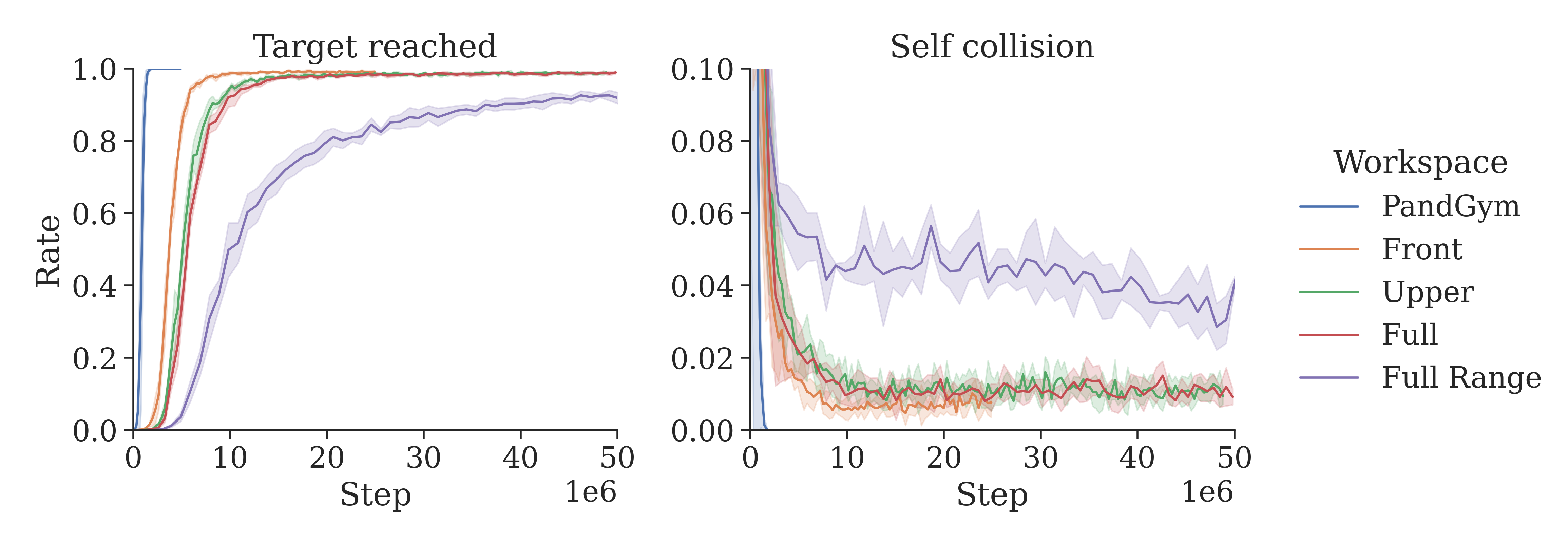}
    \caption{Training curves for the UR5e with PPO across different workspace sizes. A smaller workspace or a more constrained initialization range leads to faster learning and better overall performance.}
    \label{fig:reaching_workspace_ppo_ur5e}
\end{figure}

\begin{figure}
    \centering
    \includegraphics[width=1.0\linewidth]{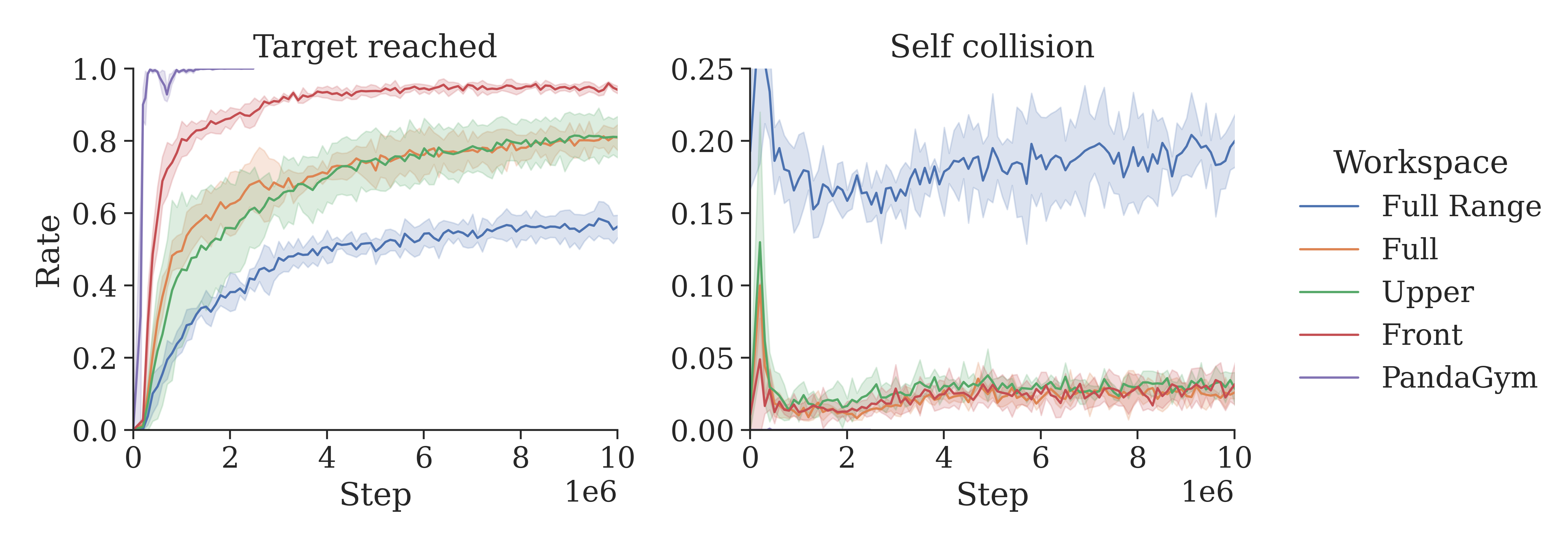}
    \caption{Training curves for the UR5e with SAC across different workspace sizes. The differences between the workspace sizes are even more pronounced with a significant penalty on overall performance for larger workspaces.}
    \label{fig:reaching_workspace_sac_ur5e}
\end{figure}

As shown in Figure \ref{fig:reaching_workspace_ppo_ur5e}, task difficulty increases with workspace size.  An agent trained in the \textit{PandaGym} workspaces achieves 100\% target reached with zero collisions in just 1.5 million steps. In contrast, an agent in the \textit{Front} workspace requires 10 million steps to reach 99\% target reached and never eliminates a residual 1\% collision rate.
This performance gap is even more pronounced when using SAC (Figure \ref{fig:reaching_workspace_sac_ur5e}).
Using the Franka Emika Robot with PPO, the difference between the \textit{PandaGym} and \textit{Front} workspace is similar, however, the between the three larger workspaces (\textit{Front}, \textit{Upper}, \textit{Full}) the differences are less pronounced.  

Our cross-evaluation analysis reveals that policies trained in smaller workspaces do not generalize to larger ones (Table \ref{tab:reaching_workspace_generalization}). For instance, the \textit{PandaGym} agent's performance collapses when evaluated in the \textit{Front} or \textit{Full} workspaces. Similarly, the agent trained in the \textit{Front} workspace experiences a significant performance drop to 64.5\% when evaluated on the \textit{Full} workspace. 
Conversely, policies trained in larger workspaces generalize perfectly to smaller ones. 
Furthermore, we observed that the generalization is also sensitive to the initial joint distribution. A policy trained in the \textit{Full} workspace with a limited joint range on initialization saw its target reached rate drop from 98.9\% to 60.1\% with an increase from 1\% to 17.8\% in self-collisions when evaluated with a full-range (\textit{Full Range}) initialization.
This indicates that the initialization of the robot's joint positions is crucial to include the full range and distribute the poses evenly over the workspace. Methods like using the final pose from the previous episode as the initial pose for the next episode \cite{kumar2021joint} might lead to poor generalization. 
\begin{table}
    \centering
    \caption{Generalization performance of agents trained on one workspace and evaluated on another using the UR5e robot and PPO algorithm. Larger workspaces generalize to smaller ones, but not vice versa.}
    \resizebox{0.485\textwidth}{!}{\begin{tabular}{rrccc}
        \hline\hline
        \textbf{\begin{tabular}[c]{@{}c@{}}Training \\ workspace\end{tabular}} & 
        \textbf{\begin{tabular}[c]{@{}c@{}}Evaluation \\ workspace\end{tabular}} & 
        \textbf{\begin{tabular}[c]{@{}c@{}}Target \\ reached\end{tabular}} & 
        \textbf{\begin{tabular}[c]{@{}c@{}}Self \\ collision\end{tabular}} & 
        \textbf{\begin{tabular}[c]{@{}c@{}}Distance \\ to target\end{tabular}} \\
        \hline
        PandaGym   & PandaGym   & 1.000 ± 0.000 & 0.000 ± 0.000 & 0.001m ± 0.000 \\
        Front      & Front      & 0.993 ± 0.001 & 0.007 ± 0.001 & 0.002m ± 0.000 \\
        Upper      & Upper      & 0.988 ± 0.002 & 0.010 ± 0.002 & 0.002m ± 0.000 \\
        Full       & Full       & 0.989 ± 0.001 & 0.010 ± 0.002 & 0.002m ± 0.000 \\
        Full Range & Full Range & 0.959 ± 0.003 & 0.025 ± 0.003 & 0.005m ± 0.001 \\
        \hline
        PandaGym   & Front      & 0.139 ± 0.035 & 0.143 ± 0.066 & 0.211 ± 0.038 \\
        PandaGym   & Full       & 0.040 ± 0.016 & 0.632 ± 0.069 & 0.356 ± 0.106 \\
        Front      & Full       & 0.645 ± 0.021 & 0.186 ± 0.019 & 0.082m ± 0.005 \\
        Upper      & Full       & 0.945 ± 0.012 & 0.024 ± 0.006 & 0.007m ± 0.002 \\
        \hline
        Front      & PandaGym   & 1.000 ± 0.000 & 0.000 ± 0.000 & 0.001m ± 0.000 \\
        Full       & PandaGym   & 1.000 ± 0.000 & 0.000 ± 0.000 & 0.002m ± 0.001 \\
        Full       & Upper      & 0.988 ± 0.002 & 0.011 ± 0.002 & 0.002m ± 0.000 \\
        \hline
        Full       & Full Range &  0.601 ± 0.023 & 0.178 ± 0.027 & 0.161m ± 0.006 \\
        \hline\hline
    \end{tabular}}
    \label{tab:reaching_workspace_generalization}
\end{table}

\subsection{Reach-Avoid Task}
The reach-avoid task extends the reach task by introducing a static, randomly placed spherical obstacle requiring modifications to the observation and reward. Based on our reach task analysis and due to the page limit, we present a summary of the key findings for the reach-avoid task and do not include all detailed analysis as on the reach task above. All experiments use the relative position control. 

The best-performing PPO agent on the reach-avoid task achieves a target reached rate of \textbf{86.8\%} on the UR5e (10 cm obstacle radius, 2 cm target), with 2.7\% self collisions and 4.8\% obstacle collisions. The Franka Emika Robot performed better, achieving \textbf{95.2\%} target reached with 0.1\% self-collisions and 2.5\% obstacle collisions. 

Unlike in the reach task, the semi-sparse reward formulation proved slightly superior for the reach-avoid task. While it learned slightly slower, it converged to a safer policy with 1.5\% fewer self-collisions and 1\% fewer obstacle collisions compared to the dense reward agent. 

\begin{figure}
    \centering
    \includegraphics[width=1.0\linewidth]{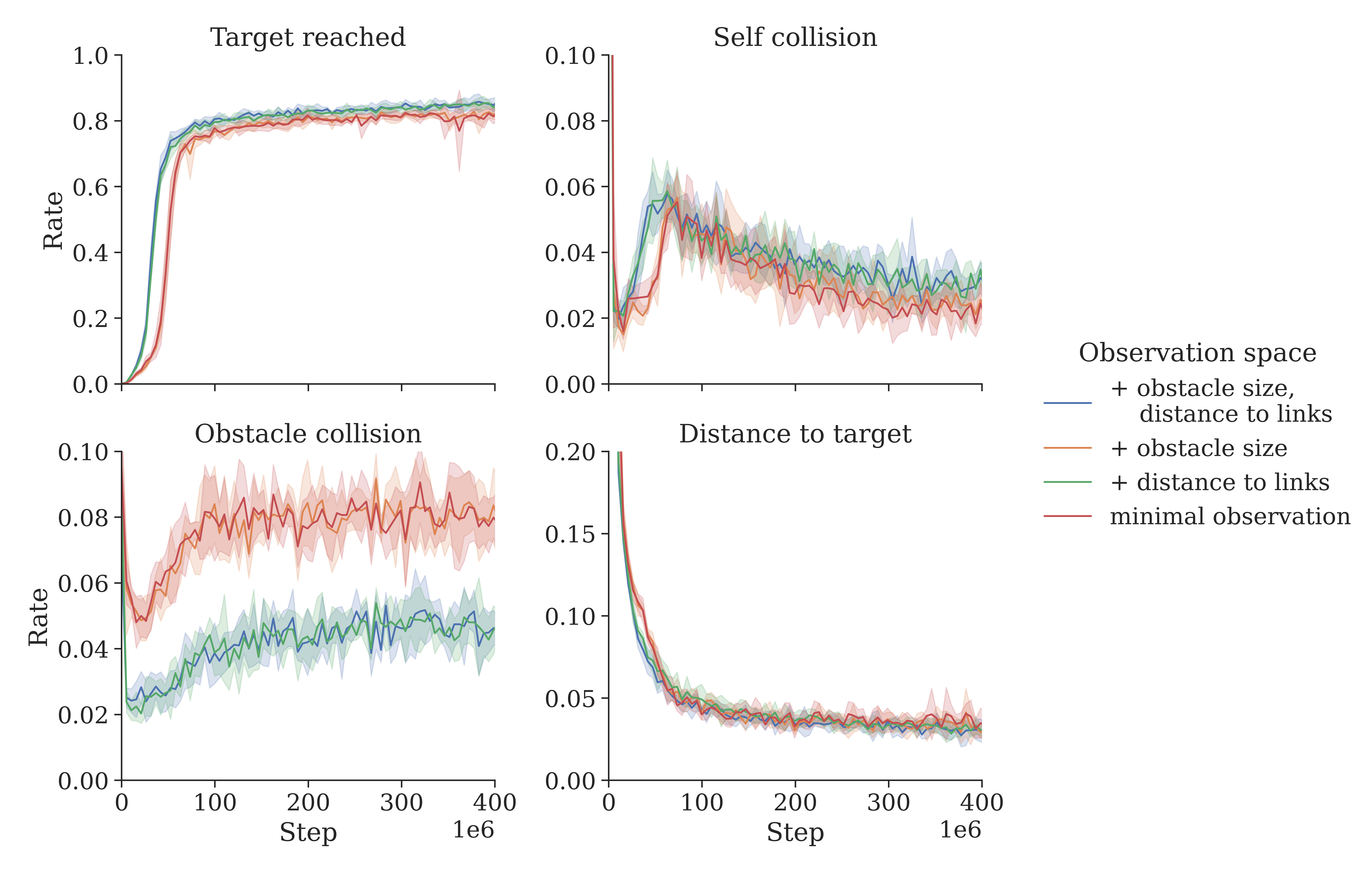}
    \caption{Comparison of different observation spaces for the reach-avoid task. Including the link to obstacle distances significantly reduces obstacle collisions.}
    \label{fig:static_observation_space}
\end{figure}

For the observation space, adding the Euclidean distance from each robot link to the obstacle was critical. As shown in Figure \ref{fig:static_observation_space}, this single feature reduced the obstacle collision rate by nearly half (from 8\% to 4.5\%). Including the obstacle's size provided no benefit, likely because its size was not randomized during training.

Due to the increased task complexity, a larger network (4×256) achieved slightly better results than the 3×256 architecture used in the reach task. Further increasing it to 5 layers offered no additional improvement.

Restricting the reach-avoid task to the smaller \textit{Front} workspace yields a significant improvement in performance compared to the full workspace. In this constrained setting, the UR5e agent achieved a 93.5\% target reached rate with 1.1\% self-collisions and 3.0\% obstacle collisions. 
The Franka Emika Robot performed even better, but with less of a performance gain over the already good results in the \textit{Full} workspace. It reached 96.1\% target reached with minimal self-collisions (0.1\%) and only 1.7\% obstacle collisions.
This result further highlights the profound impact of workspace size, demonstrating that its influence magnifies with increased task complexity. Similar to the findings in \cite{kumar2021joint} the performance penalty for introducing an obstacle in this smaller tabletop-like setting is relatively minor. While our results are comparable in its performance to those in \cite{kumar2021joint}, a direct 1-to-1 comparison is difficult due to their limited experimental setup and missing baselines.

\subsection{Influence of the Obstacle size}
\begin{figure}
    \centering
    \includegraphics[width=1.0\linewidth]{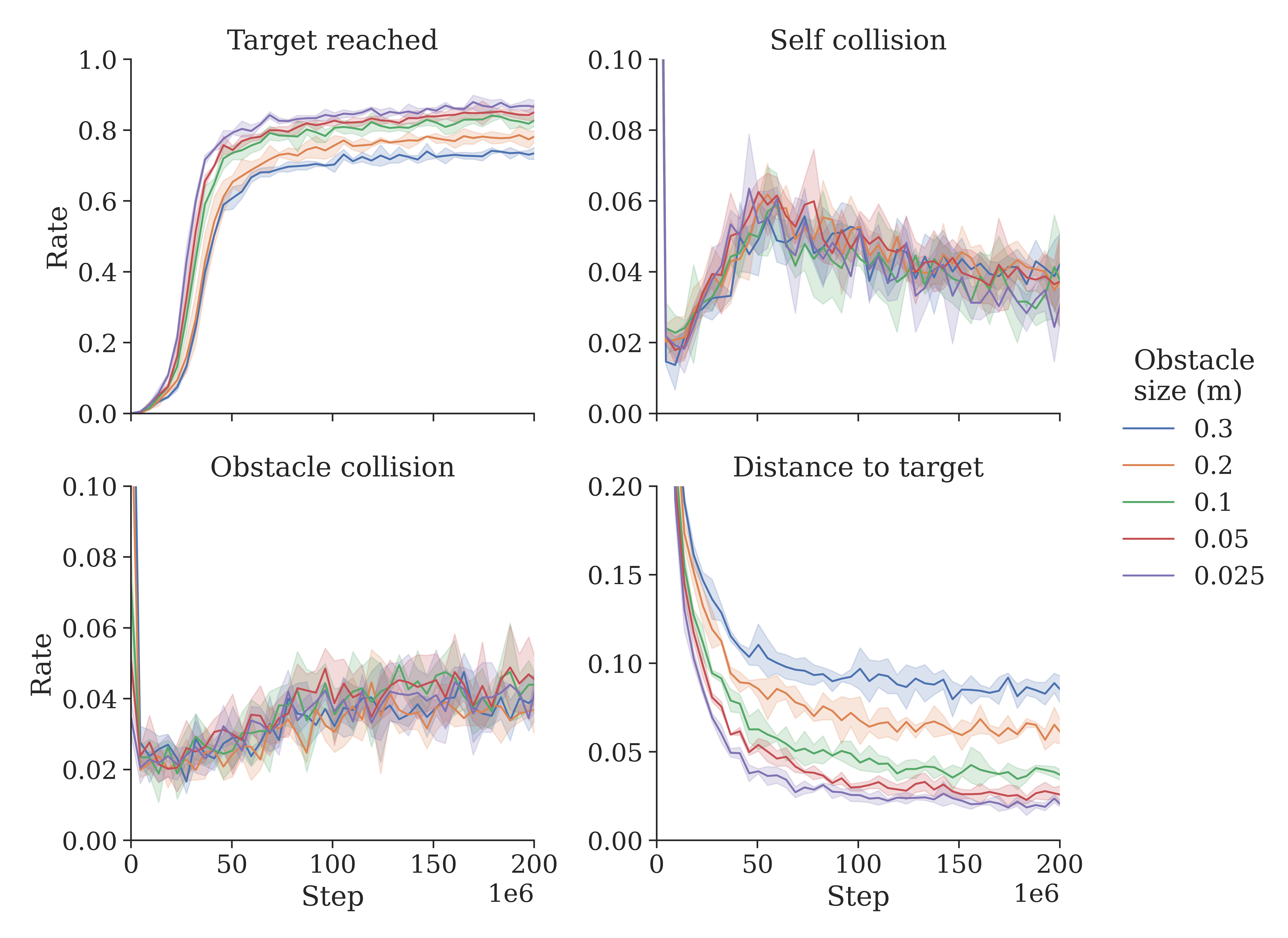}
    \caption{Training curves for the reach-avoid task with different obstacle sizes (radius). Larger obstacles lead to a lower target reached rate.}
    \label{fig:static_obstacle_size}
\end{figure}

The obstacle size notably impacts the agent's performance. As shown in Figure \ref{fig:static_obstacle_size}, while the final obstacle collision rate remains at a low residual level of approximately 4\% across different sizes, the target reached rate steadily declines as a larger obstacle is used.
This indicates that bigger obstacles increase the difficulty for the agent to find a viable path to the target. This effect is not simply due to the larger obstacle physically blocking the target, as it persists when we use a target position sampling technique that uses the Cartesian end-effector position of a randomly sampled pose of the robot and therefore guarantees reachability.

Our cross-evaluation, detailed in Table \ref{tab:static_obst_size}, reveals that the agents trained on a fixed obstacle size overfit to that specific size. The results follow an expected trend: agents trained on smaller obstacles perform poorly when evaluated against larger ones, while agents trained on larger obstacles generalize more effectively to smaller ones. 
\begin{table}
    \centering
    \caption{Performance of agents trained on a specific obstacle size and evaluated on various sizes. The \textit{reach} is an reach task agent trained without any obstacle. The size depicts the radius of the spherical obstacle.}
    \resizebox{0.485\textwidth}{!}{\begin{tabular}{ccccc}
        \hline\hline
        \textbf{\begin{tabular}[c]{@{}c@{}}Trained \\ size\end{tabular}} & 
        \textbf{\begin{tabular}[c]{@{}c@{}}Evaluated \\ size\end{tabular}} & 
        \textbf{\begin{tabular}[c]{@{}c@{}}Target \\ reached\end{tabular}} & 
        \textbf{\begin{tabular}[c]{@{}c@{}}Self \\ collision\end{tabular}} & 
        \textbf{\begin{tabular}[c]{@{}c@{}}Obstacle \\ collision\end{tabular}} \\
        \hline
        0.025 & 0.025 & 0.880 ± 0.016 & 0.037 ± 0.007 & 0.037 ± 0.002 \\
        0.05  & 0.05  & 0.860 ± 0.014 & 0.038 ± 0.005 & 0.040 ± 0.002 \\
        0.1   & 0.1   & 0.826 ± 0.027 & 0.043 ± 0.005 & 0.041 ± 0.002 \\
        0.2   & 0.2   & 0.796 ± 0.015 & 0.047 ± 0.006 & 0.040 ± 0.007 \\
        0.3   & 0.3   & 0.781 ± 0.017 & 0.049 ± 0.004 & 0.042 ± 0.002 \\
        \hline
        0.025 & 0.1   & 0.844 ± 0.015 & 0.038 ± 0.008 & 0.066 ± 0.002 \\
        0.05  & 0.1   & 0.842 ± 0.014 & 0.040 ± 0.006 & 0.052 ± 0.002 \\
        0.2   & 0.1   & 0.814 ± 0.017 & 0.046 ± 0.006 & 0.037 ± 0.004 \\
        0.3   & 0.1   & 0.803 ± 0.022 & 0.047 ± 0.004 & 0.033 ± 0.002 \\
        \hline
        reach & 0.025 & 0.844 ± 0.007 & 0.021 ± 0.002 & 0.125 ± 0.005 \\
        reach & 0.05  & 0.813 ± 0.009 & 0.020 ± 0.002 & 0.156 ± 0.006 \\
        reach & 0.1   & 0.758 ± 0.009 & 0.019 ± 0.001 & 0.213 ± 0.007 \\
        reach & 0.2   & 0.673 ± 0.006 & 0.018 ± 0.001 & 0.301 ± 0.006 \\
        reach & 0.3   & 0.603 ± 0.006 & 0.016 ± 0.002 & 0.373 ± 0.006 \\
        \hline\hline
    \end{tabular}}
    \label{tab:static_obst_size}
\end{table}

\section{Discussion}
In this paper, we introduced a comprehensive simulation benchmark for evaluating DRL agents on reach and reach-avoid tasks in complex realistic settings. We conducted a thorough analysis ranging from a simple reach task in a limited workspace to challenging reach-avoid settings, using the robot's full workspace.
Our high-speed, parallelized simulation framework enabled, for the first time, successful training in these realistic and demanding environmental settings. We achieved state-of-the-art results with success rates of 96.1\% (UR5e) and 98.8\% (Franka Emika Robot) for the reach task and 86.8\% (UR5e) and 95.2\% (Franka) for the reach-avoid task. 

Our results confirm that while DRL agents can achieve near-perfect performance in constrained settings like PandaGym, aligning with previous studies, performance degrades and convergence slows significantly in more complex, full-workspace scenarios. We conclude that agent performance depends heavily not only on the learning algorithm but also on environmental factors, particularly the workspace size and the initial pose distribution. 
Therefore, we assert that it is critical for future research to thoroughly document these experimental conditions to ensure reproducibility and comparable results.

This study is not without limitations, including the inherent sim-to-real gap and the restriction to static spherical obstacles. Future work should prioritize bridging this gap by extending the benchmark to physical robotic platforms and incorporating realistic noise models. 
Further expansion could involve introducing dynamic obstacles and more complex obstacle geometries to push the boundaries of DRL and robotic manipulation. 

\addtolength{\textheight}{-12cm}   



\bibliographystyle{IEEEtran}
\bibliography{IEEEtranBST/IEEEabrv, bibliography}
\end{document}